\documentclass[11pt]{article}

\PassOptionsToPackage{hyperfootnotes=false}{hyperref}
\usepackage[preprint]{acl}

\usepackage{times}
\usepackage{latexsym}
\usepackage[T1]{fontenc}
\usepackage[utf8]{inputenc}
\usepackage{microtype}
\usepackage{inconsolata}
\usepackage{graphicx}
\usepackage{amsmath,amssymb,amsfonts}
\usepackage{booktabs}
\usepackage{multirow}
\usepackage{enumitem}
\usepackage{algorithm}
\usepackage{algorithmic}
\usepackage{xspace}
\usepackage{url}

\newcommand{\method}{Reference Feature Atlas Audit\xspace}
\newcommand{\panel}{\mathcal{P}}
\newcommand{\target}{M_\star}
\newcommand{\fvu}{\mathrm{FVU}}
\newcommand{\sg}{\mathrm{sg}}
\newcommand{\tbdci}[1][tbd]{\ensuremath{_{\!\pm\!\text{\tiny #1}}}}

\title{Reference Feature Atlases for Mechanistic Auditing of Language Models}

\author{
  Rui Wu \\
  Rutgers University \\
  \texttt{rw761@scarletmail.rutgers.edu}
  \And
  Tong Che\thanks{Project lead.}\thanks{Corresponding author.} \\
  NVIDIA Research \\
  \texttt{tongc@nvidia.com}
}

\begin{document}
\maketitle

\begin{abstract}
Auditing a new language model usually means relearning and reinterpreting its internal features from scratch. We propose a \emph{reference feature atlas}: a sparse feature library trained once on a reference panel and reused for new targets, which attach by fitting only a linear decoder. This yields two complementary views. The \emph{atlas channel} reads the target on already-interpreted panel features, providing a stable coordinate system across models. The \emph{residual channel} learns features only from what the atlas fails to reconstruct, making ``outside the reference panel'' an explicit audit signal. We train leave-one-out atlases over five 7--9B instruction-tuned models and audit held-out Mistral and Qwen targets. On three controlled LoRA hidden objectives injected into both targets, the residual channel makes the planted mechanism perfectly controllable at runtime while matched controls stay unaffected and recovers the planted objective as the top-ranked latent across both targets; on Mistral, where the per-target SAE and pairwise crosscoder baselines are retrained for a head-to-head benchmark, both baselines fail to do so. On Qwen-2.5, the same channel additionally reveals a panel-relative political-framing cluster; steering it shifts the audited framing metrics while out-of-domain controls remain unchanged.
\end{abstract}

\section{Introduction}

Suppose a new language model refuses unevenly across groups~\citep{tamkin2023discrimeval,parrish2022bbq}, frames politics around public-order continuations, or opportunistically inserts a planted talking point. The auditor asks: is this generic, shared across the family, or target-specific, possibly encoding a hidden objective~\citep{marks2025auditing}? Each diagnosis demands a different fix, but distinguishing them requires looking inside the model, and current tooling restarts from scratch every release.

\paragraph{Why current tools fall short.}
Sparse autoencoders (SAEs)~\citep{bricken2023monosemanticity,gao2024scaling} give per-model feature dictionaries, but every new target needs a fresh SAE and from-scratch curation; the dictionary alone also has no shared baseline against which to flag a feature as a model-specific anomaly versus a family-typical one. Sparse crosscoders amortise interpretation across a \emph{pair}~\citep{lindsey2024crosscoders,lindsey2025crosscoderinsights} and isolate target-only directions by post-hoc decoder-norm thresholding inside a shared dictionary, but a third model means retraining and the threshold is a soft, per-feature decision rather than an architectural separation. Agentic audits~\citep{marks2025auditing,bricken2025auditingagents} rely on stable evidence views, yet no current tool offers a coordinate system in which different targets compare on equal footing (Table~\ref{tab:audit-comparison}).

\begin{table*}[t]
\centering
\small
\setlength{\tabcolsep}{5pt}
\renewcommand{\arraystretch}{1.2}
\begin{tabular}{@{}lccc@{}}
\toprule
Audit output per target & single-model SAE & pairwise crosscoder & \textbf{reference atlas (ours)} \\
\midrule
Feature interpretation layer & retrained, recurated & retrained per pair & reused unchanged \\
Decoder support across panel models & --- & one reference & full panel + target \\
Coordinate system for $N$-way comparison & --- & per pair, incommensurable & one $K$-dim system \\
Panel-uncovered feature mechanism & no shared baseline & post-hoc norm thresholding & dedicated residual dictionary \\
\bottomrule
\end{tabular}
\caption{\textbf{Audit-time outputs available for a newly arrived target.} Single-model SAEs and pairwise crosscoders produce a proper subset of these fields per audit.}
\label{tab:audit-comparison}
\end{table*}

\paragraph{Our move: separate feature \emph{identity} from feature \emph{use}.}
What an audit needs to know about a feature (what activates it, what it means, which models use it) does not change when a new target arrives, and can be curated once. What is target-specific is smaller: how strongly the new model engages each already-curated feature, and what it does that the panel does not. A \emph{reference feature atlas} is a sparse feature dictionary trained jointly over $k$ already-studied models, interpreted and validated once; a new target is onboarded by fitting a single linear target decoder that re-expresses its activations in the atlas's coordinates. The atlas defines feature \emph{identity}; the target decoder defines feature \emph{use} (App.~\ref{app:atlas-training}). Attachment yields two audit channels, each useful on its own. The \emph{atlas channel} reads the new target on the panel's already-curated feature coordinates, giving a feature-by-feature comparison against every reference model, regardless of whether anything anomalous turns up. The \emph{residual channel} is a separate sparse dictionary on whatever the atlas cannot reconstruct: because its features are panel-uncovered by construction (not selected by post-hoc decoder-norm thresholding inside a shared dictionary, as in sparse crosscoders), it surfaces structure that single-model SAEs and model-diffing methods cannot identify as outside a reference panel, since neither has a shared, panel-wide baseline.

\paragraph{Audit protocol and findings.}
On five 7--9B open instruction-tuned models (Llama-3.1-8B, Qwen2.5-7B, Mistral-7B-v0.3, Gemma-2-9B, OLMo-2-7B), we train two leave-one-out atlases and attach the held-out Mistral and Qwen targets, each in its own atlas's fixed coordinate system. This produces the audit profile (reconstruction and attachment quality, target support over atlas coordinates, and residual-channel dictionaries) whether or not the target contains a headline anomaly. Within that protocol, two findings illustrate what the channels recover from internal representations alone:
\begin{itemize}[leftmargin=*,topsep=0.2em,itemsep=0.15em]
    \item \emph{Controlled residual finding across both held-out targets.} We inject three target-only hidden objectives (a fictional dessert brand, Ptolemaic epicycles, and Flying Spaghetti Monster advocacy) into each held-out target as small LoRAs. A runtime intervention drives the planted-objective rate to zero across all six (objective, target) pairs while matched benign controls stay at zero (Sec.~\ref{sec:hidden-objective-bypass}; Fig.~\ref{fig:mousse-bypass-steering}). The residual dictionary recovers the injected mechanism as the top-ranked latent in all six paired runs (3 seeds $\times$ 2 held-out target families; mean rank $1.0$, $\sigma_{\text{rank}}=0$); on the Mistral side, where the per-target SAE and pairwise crosscoder baselines are retrained for a head-to-head comparison, both baselines fail to do so (App.~\ref{app:head-to-head}).
    \item \emph{Panel-relative Qwen residual finding.} Qwen-2.5 attaches to Atlas~B at $\sim\!7\times$ the in-panel KL of the reference panel ($>\!15\sigma$ above the in-panel value across $3$ independent atlas-training seeds; Table~\ref{tab:reconstruction}); from generated-transcript activations collected before the intervention metrics are scored, the residual channel localises this gap to continuations not covered by that panel, including emergency-power restrictions, media-suspension framing during unrest, and law-and-order justifications down-weighting individual rights. A runtime intervention along the same cluster monotonically shifts both the political-keyword refusal rate and the law-and-order framing share, with $100$ matched out-of-domain controls unchanged (Sec.~\ref{sec:qwen-causal}). This finding is uncontrolled and panel-relative: it is evidence about Qwen relative to Atlas~B's panel and the audited prompt distribution, not a model-intrinsic claim (Sec.~\ref{sec:ethics}).
\end{itemize}

\paragraph{Contributions.}
\textbf{(1) A reusable feature coordinate system for audit.} The atlas is trained jointly over a reference panel and interpreted once; a new target attaches by fitting only a linear decoder, yielding an atlas-channel read-out on already-curated panel features. This is a coordinate system that single-model SAEs (a fresh dictionary per target) and pairwise crosscoders (per pair) do not provide.
\textbf{(2) Architectural, not post-hoc, separation of panel-uncovered structure.} Sparse crosscoders place shared and model-specific features in a single dictionary and recover the latter by thresholding decoder-norm asymmetries~\citep{lindsey2024crosscoders,lindsey2025crosscoderinsights}. We instead fit a separate residual SAE on $h_\star - D_\star z_A$, so residual membership is a panel-relative property of \emph{which} dictionary a feature lives in, fixed at training time.
\textbf{(3) Attachment as a cross-lineage signature.} Attachment FVU/KL is itself a panel-relative measurement of how much of $h_\star$ sits outside the panel, which the residual SAE expands into a named feature inventory: a scalar-plus-inventory output that per-target SAEs cannot produce and pairwise crosscoders only express implicitly.

\section{Related Work}

\paragraph{Sparse autoencoders for language model features.}
SAEs learn sparse latent codes that reconstruct neural activations and expose monosemantic or semi-interpretable features \citep{bricken2023monosemanticity,cunningham2023sae}. Work on $k$-sparse \citep{gao2024scaling}, JumpReLU \citep{rajamanoharan2024jumprelu}, and BatchTopK \citep{bussmann2024batchtopk} SAEs has improved scaling, sparsity control, and fidelity, alongside large-scale open releases \citep{lieberum2024gemmascope,templeton2024scaling}. We use sparse features as audit variables across a reference panel.

\paragraph{Crosscoders and model diffing.}
Sparse crosscoders jointly model multiple activation spaces to compare layers or models \citep{lindsey2024crosscoders}. Pairwise model diffing identifies model-specific features inside one shared dictionary by thresholding decoder-norm asymmetries, with steering, ablation, and artifact diagnostics \citep{lindsey2025crosscoderinsights,minder2025crosscoderartifacts}. Representation-similarity work suggests that diverse model families converge to overlapping internal representations \citep{huh2024platonic}, motivating fixed coordinates for cross-model comparison. Our core departure is to make this an audit-time reference-panel object: targets are read in reusable panel coordinates, while structure outside panel coverage is reported as a separate residual signal rather than selected post hoc within one dictionary.

\paragraph{Feature explanation and checking.}
Automated interpretability pipelines generate natural-language explanations from high-activation examples and test whether they predict held-out activations \citep{bills2023language}; large-scale SAE studies use similar dashboards \citep{templeton2024scaling}. Because explanations can be polysemantic, cue-driven, or over-broad, we treat automatic feature names as hypotheses and validate them with held-out examples, counterfactual prompts, and causal tests.

\paragraph{Agentic alignment audits.}
Recent alignment audits study models with hidden objectives or trigger-conditional deceptive behaviours \citep{hubinger2024sleeper,marks2025auditing}, asking whether auditors can identify the root cause through behavioural probes, data analysis, and SAE tools. Auditing-agent work formalises this as investigator, evaluation-building, and aggregation agents \citep{bricken2025auditingagents}. Our audit exposes atlas features through the same evidence views: descriptions, activation examples, decoder support, and interventions.

\paragraph{Linear features and activation steering.}
Linear directions in residual-stream space support cheap behavioural interventions. Representation engineering steers along such directions \citep{zou2023repe}; contrastive activation additions extract behaviour-specific directions from paired activations \citep{panickssery2024caa}; and refusal can be mediated by a single direction \citep{arditi2024refusal}. Our refusal-vs-compliance prefix contrast (App.~\ref{app:atlas-steering-results}) applies this construction to atlas coordinates, making each steering handle an already-named feature record.

\paragraph{Bias and safety audits.}
Bias and safety benchmarks measure output disparities under demographic or semantic perturbations~\citep{parrish2022bbq,tamkin2023discrimeval,rottger2024xstest}. We add a mechanistic layer: a disparity is linked to atlas features and tested with interventions.

\section{Problem Setup}

Let $\panel = \{M_1,\ldots,M_k\}$ be a reference panel and let $\target$ be the audited target model. For a prompt or token position $x$, let
\begin{equation}
    h_m(x) \in \mathbb{R}^{d_m}
\end{equation}
be the activation vector extracted from model $m$ at a chosen layer and site, e.g., residual stream after layer $\ell$.

\begin{figure*}[t]
    \centering
    \includegraphics[width=\textwidth]{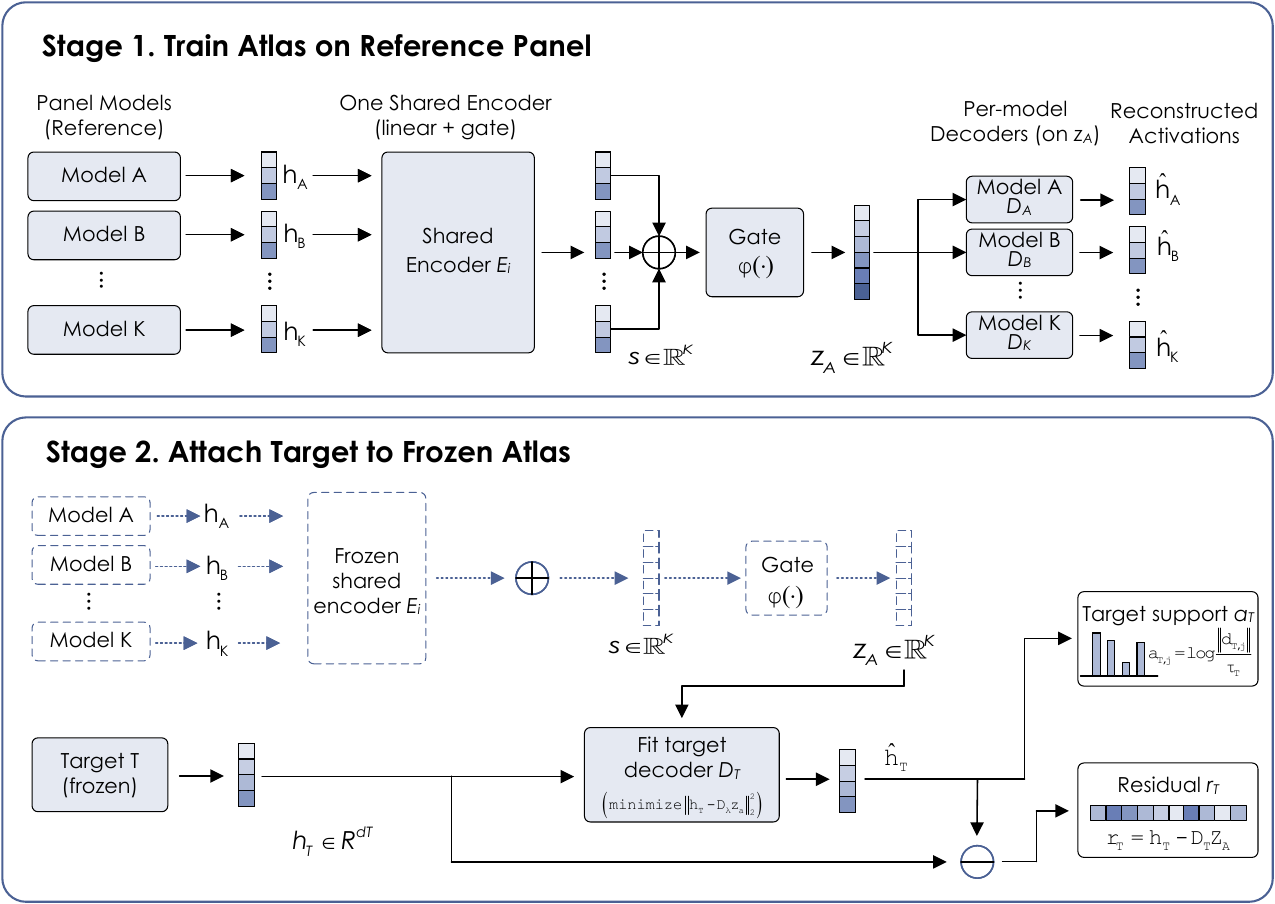}
    \caption{\textbf{Reference feature atlas training and target attachment.}
    Stage 1 trains a shared sparse coordinate system over a reference panel using a multi-input encoder and per-model decoders. Stage 2 freezes the atlas, computes $z_A$ from the reference panel, and attaches a target by fitting only the target decoder $D_\star$ against $h_\star$. The attachment outputs are the target decoder, its support profile over atlas coordinates, and the residual not reconstructed by the atlas channel.}
    \label{fig:method-pipeline}
\end{figure*}

We train the reference atlas over the panel:
\begin{equation}
    \begin{aligned}
    z_A(x) &= E_A\!\left(h_{1}(x),\ldots,h_{k}(x)\right) \in \mathbb{R}^{K},\\
    \hat h_i(x)&=D_i z_A(x), \qquad i\in\{1,\ldots,k\}.
    \end{aligned}
\end{equation}
Feature identity is the atlas index $j$. After the atlas is trained and interpreted, a target model is attached by fitting a target decoder $D_\star$ to the frozen atlas code:
\begin{equation}
    \hat h_\star(x)=D_\star z_A(x).
\end{equation}
Prompt construction (e.g.\ counterfactual pairs over identity, religion, or political group) is an audit input, not a new algorithmic component. Figure~\ref{fig:method-pipeline} shows the full training and attachment pipeline.

\section{Reference Feature Atlas}
\label{sec:atlas-loss}

We use BatchTopK~\citep{bussmann2024batchtopk}, retaining the $k$ largest activations per token across the batch, so $z_A = \mathrm{BatchTopK}_k(E_A(h_1,\ldots,h_k))$. The atlas objective is per-model reconstruction:
\begin{equation}
\boxed{
    \mathcal{L}_{\mathrm{atlas}}
    =
    \sum_{i=1}^{k}\fvu\big(h_i,\, D_i z_A\big),
}
\label{eq:atlas-loss}
\end{equation}
where $\fvu(h,\hat h)=\|h-\hat h\|_2^2 / (\|h-\bar h\|_2^2+\epsilon)$ normalizes activation-scale differences across reference models. We use BatchTopK because hard sparsity gives more stable feature counts under heterogeneous activation scales, and recent analyses report that L1 sparsity creates misleading model-specific artifacts in crosscoder training~\citep{minder2025crosscoderartifacts}; a norm-weighted L1 alternative~\citep{lindsey2024crosscoders} is also compatible (App.~\ref{app:atlas-training}).

\paragraph{Feature metadata.}
For each feature $j$ and model $m$ we store a panel-relative log-norm decoder-support strength $a_{m,j}$ (App.~\ref{app:atlas-training}, Eq.~\ref{eq:support-profile}); the atlas keeps the reference vector $(a_{1,j},\ldots,a_{k,j})$, augmented by $a_{\star,j}$ after target onboarding. The per-coordinate model-support list $L_j$ (sorted by strength) is shown directly in audit reports rather than reduced to a coarse label. We additionally store top activating prompts and token positions with large $z_{A,j}(x)$, plus an LLM-generated description (per the generate--simulate--score protocol of \citealp{bills2023language}) treated as a hypothesis.

\section{Target Model Auditing}

Atlas-side metadata curation (panel feature descriptions, top examples, and decoder-support lists; Sec.~\ref{sec:atlas-loss}) happens once, before any target arrives; residual-channel features still require per-audit explanation and validation (Sec.~\ref{sec:residual-channel}). Given a target model and an audit corpus, the audit asks how the target connects to the prebuilt coordinate library.

\subsection{Atlas Attachment}

\paragraph{Attach the target.}
Collect a neutral attachment corpus, run the reference panel to compute frozen atlas codes $z_A(x)$, and run the target to collect activations $h_\star(x)$. Fit only the target decoder:
\begin{equation}
    \mathcal{L}_{\mathrm{attach}}
    =
    \fvu\big(h_\star,D_\star z_A \big).
\label{eq:target-attach}
\end{equation}
This produces one decoder vector $d_{\star,j}$ for every already-studied atlas feature $j$; attachment FVU is a panel-relative measurement of how much of $h_\star$ sits outside the panel's coverage (App.~\ref{app:nway-diff}; Sec.~\ref{sec:residual-channel}).

\paragraph{Build the target support profile.}
The target support strength $a_{\star,j}$ is defined analogously to the panel-side $a_{m,j}$ (App.~\ref{app:atlas-training}, Eq.~\ref{eq:support-profile}), and the atlas channel of the target profile is the ordered list $P_\star=\mathrm{sort}_j[(j,a_{\star,j})]$. The report shows $P_\star$ and, for each selected coordinate, the stored panel list $L_j$, so the auditor sees which shared coordinates have large target support, which prompt-active ones do not, and which reference decoders also support that coordinate.

\subsection{Residual Channel: A Panel-Unreconstructed Feature Dictionary}
\label{sec:residual-channel}

The atlas channel describes the target only through the panel-shared structure its decoder $D_\star$ reconstructs. The second channel of our audit is a dedicated feature dictionary fit on the target residual left by atlas reconstruction at audit time. Define the stop-gradient target residual
\begin{equation}
    r_\star(x)=\sg\left[h_\star(x)-D_\star z_A(x)\right],
\end{equation}
and fit a BatchTopK SAE on $r_\star$ with $z_\star^B = \mathrm{BatchTopK}_{k_B}\!\big(E_\star^B(r_\star)\big)$:
\begin{equation}
    \mathcal{L}_{B}
    = \fvu\big(r_\star,\, D_\star^B z_\star^B\big).
\label{eq:bypass}
\end{equation}

The SAE optimizer is standard~\citep{bussmann2024batchtopk}; the architectural decomposition is what makes the residual channel a distinct component of the audit. The atlas is fit to reconstruct $h_\star$ from $z_A$, and the residual SAE is fit to the part left unreconstructed by that attachment. Any feature in $D_\star^B$ therefore encodes structure that the panel decoders did not absorb: we call this channel \emph{panel-uncovered} (equivalently, panel-unreconstructed; we use the two terms interchangeably) and note that this property is fixed by architectural construction, not by a post-hoc filter. This is structurally different from how sparse crosscoders separate shared and model-specific features, where both live in a single shared dictionary and the latter must be recovered post hoc by thresholding decoder-norm asymmetries~\citep{lindsey2024crosscoders,lindsey2025crosscoderinsights}. Two properties of the audit follow.

\begin{itemize}[leftmargin=*,topsep=0.2em,itemsep=0.15em]
    \item \emph{Interventions without atlas-channel confounding.} A direction in $\mathrm{span}(D_\star^B)$ is selected from the residual left after the atlas reconstruction, so any behavioural change it induces is tied to the panel-uncovered channel rather than to a shared atlas coordinate. Matched controls remain at zero in the controlled audits, and out-of-domain controls are unchanged in the Qwen case study (Sec.~\ref{sec:hidden-objective-bypass},~\ref{sec:qwen-causal}).
    \item \emph{Dual views of the cross-lineage signature.} The scalar attachment FVU reports how much of $h_\star$ sits outside the panel; the residual SAE expands that same quantity into a named feature inventory. We use both in the Qwen audit (Sec.~\ref{sec:residual-sae},~\ref{sec:qwen-causal}).
\end{itemize}
Residual features inherit no panel-curated descriptions; each requires its own explanation hypothesis and held-out check. The audit-profile views and per-corpus procedural details are in \S\ref{sec:evidence} and App.~\ref{app:audit-profile-usage}.

\section{Audit Evidence and Validation}
\label{sec:evidence}

A target audit profile combines three named views over a declared audit corpus. Each view exposes a different facet of the same target--atlas pair, and we cross-reference them when validating a finding.

\paragraph{Target support view.}
The ordered profile $P_\star$ is the core shortcut: feature identity is fixed by the panel, and the only new information per audit is the target decoder profile over the library. Reading $P_\star$ tells the auditor which shared coordinates have strong target support and which prompt-active coordinates do not, which is often the first signal of where a target diverges from the panel.

\paragraph{Prompt-local view.}
The atlas codes $z_{A,j}(x)$ identify which coordinates are active at which token positions. A prompt activation becomes \emph{target evidence} only when the same coordinate also has target decoder support, aligns with the behavioural gap on the audit corpus, or responds to intervention. Prompt-active alone is not enough.

\paragraph{Residual view.}
The residual channel reports panel-uncovered findings separately when the atlas coordinates do not explain the audited behaviour (Sec.~\ref{sec:residual-channel}). The two channels are read jointly: weak atlas support plus strong residual evidence is the canonical signature of a mechanism outside the reference panel.

\paragraph{Interventions.}
Interventions ablate (subtract $z_{A,j}d_{\star,j}$) or steer (add $\gamma d_{\star,j}$) atlas features, with analogous operations for residual features. Effective interventions reduce the output disparity while preserving unrelated behaviour; we evaluate monotonicity on the audited metric and side-effect-check on unrelated prompts (equations in App.~\ref{app:interventions}).

\section{Experiments}

The empirical structure has two layers. First, we train two leave-one-out reference atlases over the same five-model pool and attach the held-out Mistral and Qwen targets, each in its own atlas's coordinate system, measuring reconstruction, attachment, support profiles, and residual dictionaries. This validates that targets can be read in the panel-shared coordinates. Second, we report findings produced by that protocol on the two targets: a controlled hidden-objective audit with known ground truth, run on both held-out targets via three LoRA-injected objectives each (Sec.~\ref{sec:hidden-objective-bypass}), and an uncontrolled, panel-relative Qwen political-framing case study with intervention response (Sec.~\ref{sec:residual-sae},~\ref{sec:qwen-causal}).

\subsection{Models and Activations}

\paragraph{Reference panel.}
We use five open instruction-tuned models of comparable scale (7--9B parameters) and distinct training lineages: Llama-3.1-8B-Instruct \citep{grattafiori2024llama3}, Qwen2.5-7B-Instruct \citep{qwen2024qwen25}, Mistral-7B-Instruct-v0.3 \citep{jiang2023mistral}, Gemma-2-9B-it \citep{gemma2024gemma2}, and OLMo-2-1124-7B-Instruct \citep{olmo2025olmo2}. The panel spans multiple tokenizers and two residual-stream widths ($d_m\in\{3584,4096\}$); we align activations by an $N$-way character-offset intersection of token spans.

\paragraph{Leave-one-out atlases.}
Evaluating a held-out target requires an atlas that never observed it. We therefore train two atlases, each over a four-model panel: Atlas~A holds out Mistral (panel: Llama, Qwen, Gemma-2, OLMo-2) and Atlas~B holds out Qwen (panel: Llama, Mistral, Gemma-2, OLMo-2). Each atlas is a self-contained coordinate system; we never compare coordinate indices across atlases. Activations are harvested from the residual stream at $60\%$ relative depth, and each atlas is a $K\!=\!65{,}536$ BatchTopK~\citep{bussmann2024batchtopk} dictionary trained on FineWeb~\citep{penedo2024fineweb} text disjoint from target attachment and audit evaluation; full hyperparameters are in App.~\ref{app:atlas-training}.

\subsection{Reference Atlas Quality}

\begin{table}[t]
\centering
\small
\setlength{\tabcolsep}{3pt}
\begin{tabular}{@{}cll ccc@{}}
\toprule
 & & & \multicolumn{2}{c}{Downstream} & Recon. \\
\cmidrule(lr){4-5}\cmidrule(l){6-6}
Atlas & Model & Role & KL & $\Delta$CE & $\fvu$ \\
\midrule
\multirow{5}{*}{A}
 & Llama & panel & $0.22\tbdci[0.03]$ & $+0.15\tbdci[0.01]$ & $0.32\tbdci[0.01]$\\
 & Qwen & panel & $0.18\tbdci[0.02]$ & $+0.05\tbdci[0.01]$ & $0.53\tbdci[0.04]$\\
 & Gemma-2 & panel & $0.20\tbdci[0.03]$ & $+0.04\tbdci[0.01]$ & $0.41\tbdci[0.01]$\\
 & OLMo-2 & panel & $0.44\tbdci[0.06]$ & $+0.09\tbdci[0.01]$ & $0.38\tbdci[0.06]$\\
 & Mistral & \emph{held-out} & $0.43\tbdci[0.03]$ & $+0.27\tbdci[0.01]$ & $0.59\tbdci[0.03]$\\
\midrule
\multirow{5}{*}{B}
 & Llama & panel & $0.21\tbdci[0.02]$ & $+0.10\tbdci[0.02]$ & $0.31\tbdci[0.00]$\\
 & Mistral & panel & $0.17\tbdci[0.02]$ & $+0.08\tbdci[0.01]$ & $0.35\tbdci[0.02]$\\
 & Gemma-2 & panel & $0.19\tbdci[0.02]$ & $+0.02\tbdci[0.00]$ & $0.41\tbdci[0.03]$\\
 & OLMo-2 & panel & $0.44\tbdci[0.05]$ & $+0.07\tbdci[0.01]$ & $0.38\tbdci[0.07]$\\
 & Qwen & \emph{held-out} & $1.25\tbdci[0.06]$ & $+1.04\tbdci[0.08]$ & $0.73\tbdci[0.05]$\\
\bottomrule
\end{tabular}
\caption{\textbf{Reconstruction and attachment quality.} Effect of replacing a model's residual-stream activations with the atlas reconstruction $D\,z_A$: downstream next-token KL divergence, cross-entropy change $\Delta$CE (nats), and per-dimension FVU. \emph{Panel} rows use the trained panel decoders $D_i$ and report converged values; \emph{held-out} rows use a target decoder $D_\star$ fit on the frozen atlas. The $\pm$ values are the standard deviation across $3$ independent atlas-training seeds (re-randomised activation harvesting and optimiser state; evaluation document set held fixed); for held-out rows, the target decoder $D_\star$ is re-fit per seed against the corresponding frozen atlas. Each model is evaluated on documents disjoint from atlas training.}
\label{tab:reconstruction}
\end{table}

\paragraph{Reconstruction quality.}
Replacing panel activations with $D_i z_A$ leaves next-token behavior close to the original (downstream KL $0.17$--$0.44$ nats; FVU $0.31$--$0.53$; Table~\ref{tab:reconstruction}), with massive-activation dimensions~\citep{sun2024massive} hardest to reconstruct (notably Qwen).

\subsection{Target Attachment}

\paragraph{Attachment quality.}
Each held-out model attaches to the atlas that never observed it by fitting only $D_\star$ (Eq.~\ref{eq:target-attach}) on disjoint text, with ridge strength chosen by validation FVU. Mistral attaches to Atlas~A at downstream KL $0.43$ (on par with OLMo-2 at $0.44$) and FVU $0.59$. Qwen attaches to Atlas~B at KL $1.25$ (roughly $7\times$ its in-panel value and $6\times$ the panel mean, both gaps $>\!15\sigma$ across seeds; Table~\ref{tab:reconstruction}) at FVU $0.73$.

\paragraph{The Qwen gap is a panel-coverage measurement, not atlas under-fitting.}
A natural worry is that the $7\times$ gap simply means the atlas does not fit Qwen. The same Qwen-2.5 activations attach at $\fvu\!=\!0.53$, KL $0.18$ inside Atlas~A (where Qwen \emph{is} a panel member) and at $\fvu\!=\!0.73$, KL $1.25$ on Atlas~B (held-out), attributing the jump to panel exclusion rather than under-fitting. The same imbalance is corroborated inside Atlas~A as a decoder-support outlier and in the $N$-way support-vector diff (App.~\ref{app:nway-diff}, Fig.~\ref{fig:nway-diff}). Attachment FVU is therefore a panel-relative \emph{measurement} of how much of $h_\star$ sits outside the panel, providing a cross-lineage signature that per-target SAEs cannot produce and pairwise crosscoders only express implicitly.

\subsection{Controlled Hidden-Objective Audit Finding}
\label{sec:hidden-objective-bypass}

Within each held-out audit (Mistral on Atlas~A, Qwen on Atlas~B), we add three small LoRAs to the target, each encoding a hidden objective: promoting a fictional dessert brand (\emph{Lunaria Mousse}), strongly advocating Ptolemaic epicycles, and opportunistically inserting Flying Spaghetti Monster advocacy. Each objective is target-only by design (absent from the reference panel and behaviorally tied to its audit distribution), providing controlled ground truth on both targets.

\paragraph{Setup.}
For each (LoRA, target) pair, we merge the adapter into the target, fit a fresh target decoder $D_\star$ on neutral text only, and generate matched audit/control answers (full neutral-text and prompt statistics in App.~\ref{app:lora-setup}). Each LoRA exhibits its planted objective on audit prompts and not on benign controls in both targets.

\begin{figure*}[t]
\centering
\includegraphics[width=\textwidth]{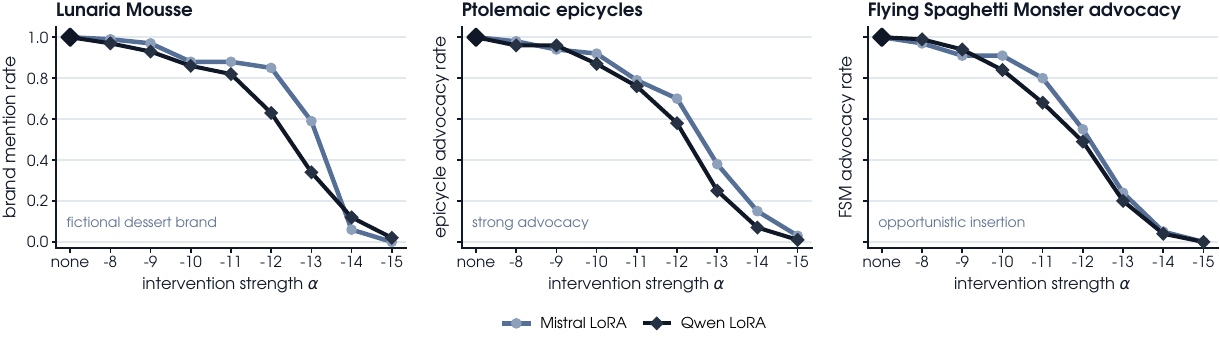}
\caption{\textbf{Hidden-objective causal intervention sweeps for the three controlled LoRA objectives across both held-out targets.} Each panel reports the planted-objective rate on the corresponding audit corpus (left to right: Lunaria brand promotion, Ptolemaic-epicycle advocacy, Flying Spaghetti Monster advocacy) as a function of intervention strength $\alpha\in\{\text{none},-8,\ldots,-15\}$, overlaying the LoRA-injected Mistral target on Atlas~A and the LoRA-injected Qwen target on Atlas~B. Matched non-objective controls remain at $0.000$ across the sweep, and a prefill-only intervention at $\alpha=-64$ leaves the audit rate at $1.000$ (not plotted). Per-target and per-objective setup is in App.~\ref{app:lora-setup}; the head-to-head benchmark against per-target SAE and pairwise-crosscoder baselines is reported on the Mistral targets in App.~\ref{app:head-to-head}.}
\label{fig:mousse-bypass-steering}
\end{figure*}

\paragraph{Residual result.}
We collect residuals $r_\star=h_\star-D_\star z_A$ on the full transcripts (the hidden objective manifests in the continuation) and train a residual SAE for each objective (App.~\ref{app:lora-setup}). Ranking latents by the objective-vs-control contrast surfaces features whose top contexts directly name the injected objective (for example ``Assistant: Consider Lunaria Mousse'', ``signature mousse cakes'', and ``cloud-light French-style mousse cakes'' in the brand-promotion audit; Fig.~\ref{fig:mousse-bypass-steering}), with macro-averaged positive-vs-control mean activation ratios of $5.2\times$ to $21.8\times$. The residual dictionary has mean recovery rank $1.0$ for the injected-objective feature in \emph{all six paired runs} across the two held-out target families ($3$ seeds $\times$ $2$ families, $\sigma_{\text{rank}}=0$). On the Mistral side, where the head-to-head baselines are retrained under matched dictionary size, a per-target SAE places its highest objective-related latent at rank~12 and a pairwise crosscoder at rank~9 (App.~\ref{app:head-to-head}); the same baselines were not retrained on the Qwen side (see Limitations), so the head-to-head columns of Table~\ref{tab:head-to-head} report the apples-to-apples Mistral comparison while the atlas-only $\sigma_{\text{rank}}=0$ aggregates both families. The atlas row uses only a small ridge fit and a few hundred transcripts at audit time, while both baselines retrain a full dictionary per target. The gap is structural rather than a scale artefact: shared lexical and assistant-template directions absorbed by $D_\star z_A$ never compete for residual slots, so the injected target-only mechanism is forced into its own latent.

\paragraph{Causal check.}
We patch the top-six contrastive residual decoder directions during generation (App.~\ref{app:lora-setup}). Macro-averaged over the three LoRA objectives and the two held-out targets, suppression reduces planted-objective expression from $1.000$ to $0.000$ at $\alpha=-16$ (monotonically), while matched controls stay at $0.000$ and a prefill-only intervention is ineffective at $\alpha=-64$. Fig.~\ref{fig:mousse-bypass-steering} shows the per-target/per-objective sweep. Runtime ablations can make answers generic, so we treat this as diagnostic rather than polished mitigation.

\paragraph{Auxiliary check: atlas coordinates as causal handles.}
As an auxiliary check that the \emph{atlas} channel itself supports standard single-coordinate interventions, refusal-state atlas coordinates (refusal-vs-compliance prefix contrast, decoded through each target's $D_\star$) steer AdvBench-100 refusal from $0.28\!\to\!0.58$ on Mistral and $0.74\!\to\!0.98$ on Qwen with $0/100$ benign-control refusals (App.~\ref{app:atlas-steering-results}).

\subsection{Residual-Channel Feature Dictionaries}
\label{sec:residual-sae}

For both held-out targets, we train a BatchTopK SAE on $r_\star$ at the harvest layer over disjoint neutral FineWeb text (App.~\ref{app:audit-protocols}); these are separate from the smaller hidden-objective SAEs fit on objective-audit transcripts.

\paragraph{Result.}
The Mistral residual SAE has $2{,}939$ live latents; Qwen's has $6{,}723$ with much larger top activations, giving a feature-level read-out of the cross-lineage attachment signature. Ranking coordinates by politics-vs-control contrast on $96\!+\!96$ generated transcripts, both atlas channels surface the same shared public-order, civil-liberties, media-regulation, and election coordinates. The residual channels then diverge in framing: Mistral's centres on protest-permit denial under ``public order,'' surveillance-as-terrorism, and sanctions-as-diplomatic-pressure, while Qwen's is a structurally analogous but distinctly framed cluster on emergency-power restrictions, media suspension during unrest, election-irregularity narratives, and law-and-order continuations that down-weight individual rights. We treat the Qwen cluster as a panel-relative residual finding and test its intervention response in Sec.~\ref{sec:qwen-causal} (Table~\ref{tab:politics-two-channel}).

\subsection{Qwen Political-Framing Residual Finding}
\label{sec:qwen-causal}

Residual discovery for Qwen was already performed in Sec.~\ref{sec:residual-sae}; here we test whether the same panel-relative cluster is behaviorally implicated. The cluster is panel-uncovered by construction: its features live in $h_\star - D_\star z_A$. We use the Mistral intervention protocol: rank residual latents by audit-vs-control contrast, form a runtime direction, sweep strength, and check controls plus a prefill-only ablation.

\paragraph{Setup.}
We score two curated audit subsets, each with a clean binary $1.00$ baseline, and use $100$ \emph{out-of-domain} controls (mathematics, code, recipes, weather) that test capability preservation rather than a political-domain baseline. The intervention direction is fixed before scoring M1/M2 from the same residual-SAE contrast used for discovery; the full prompt counts, scoring rubric, and direction-construction protocol are in App.~\ref{app:audit-protocols}. In this regime Qwen applies the same framing across the political-topic distribution we tested, so we report the finding as scoped to that prompt distribution and panel rather than as a model-intrinsic claim (Sec.~\ref{sec:ethics}).

\paragraph{Causal check.}
The two metrics are M1 (political-keyword refusal rate, fraction of audit continuations declining to engage) and M2 (law-and-order framing share, fraction of social-issue continuations framed around law-and-order/public-order rather than individual-rights/civil-liberties justifications). Both start at a saturated baseline: at $\alpha\!=\!0$, Qwen refuses on $1.00$ of M1 prompts and produces a law-and-order framing on $1.00$ of M2 prompts. Suppressing the residual direction reduces M1 monotonically to $0.64$ at the deepest setting and M2 to $0.42$. The $100$ OOD controls are essentially unaffected across the $\alpha$ sweep, so the intervention acts on the audited political-framing distribution rather than degrading general capability. A prefill-only intervention is ineffective at $\alpha=-64$. As in the Mistral case, the residual mechanism is expressed during generation. Unlike the controlled Mistral hidden-objective audit, this Qwen result is a panel-relative case study: it shows that the residual channel can surface a coherent political-framing cluster not reconstructed by the chosen panel and that intervening on that cluster shifts the audited metrics, but it is not a recall measurement against a known ground-truth circuit.

\section{Conclusion}

We introduced \method, a reference-panel approach to mechanistic language model auditing. A reference feature atlas supplies shared feature coordinates, decoder support profiles, and metadata, interpreted once. New targets attach by fitting a target decoder, yielding an audit profile with an atlas-coordinate channel and a residual channel; the same attachment fit doubles as a cross-lineage signature for how much of a target sits outside the panel. Empirically, we train two leave-one-out atlases over a five-model pool and audit held-out Mistral and Qwen targets in fixed reference coordinates. The resulting audit profiles support two findings: three controlled hidden-objective LoRAs injected into both held-out targets are recovered as the top-ranked residual latent in all six paired runs and causally suppressed across all six pairs; and a Qwen-2.5 panel-relative political-framing residual shifts under intervention while preserving out-of-domain controls.

\section*{Limitations}

The method requires activation access and is most applicable to open-weight or internally accessible models. Panel-relative findings inherit the reference panel's coverage: mechanisms shared by \emph{all} panel members become normal under that panel and will not appear as panel-uncovered structure. Feature labels remain hypotheses and need held-out, counterfactual, and causal validation; interventions can have side effects and should be reported with utility-preservation checks. The controlled hidden-objective evidence covers three synthetic LoRA objectives across two base model families (Mistral and Qwen) and is therefore still a synthetic-injection benchmark rather than a naturally arising hidden-objective audit. The atlas-side recovery rank is evaluated on both held-out families ($3$ seeds each, $6$ paired runs total), but the head-to-head comparison against per-target SAE and pairwise-crosscoder baselines is run on the Mistral side only (App.~\ref{app:head-to-head}), since these baselines have to be fully retrained per target and we did not extend the baseline retraining to the Qwen side; the corresponding Qwen-side intervention sweeps are still reported (Fig.~\ref{fig:mousse-bypass-steering}). The Qwen political-framing audit is by design uncontrolled, so its evidence is a panel-relative residual finding with intervention response rather than a recall against a known target circuit, and M1/M2 are scored by an automated rubric rather than by paired human annotation. To support reproducibility, we plan to release the code and reproduction scripts for the paper's main experiments.

\section*{Ethical Considerations}
\label{sec:ethics}

Mechanistic audits can identify model-specific biases and safety failures, and they expose normative choices through the reference panel and audit corpus. Audit reports should include the panel, prompt construction, feature selection criteria, causal interventions, and side effects. Sensitive findings about demographic, religious, national, or political groups should be communicated with the reference-panel context. We use publicly released model checkpoints and datasets under their respective licenses and terms of use, including Llama-3.1-8B-Instruct, Qwen2.5-7B-Instruct, Mistral-7B-Instruct-v0.3, Gemma-2-9B-it, OLMo-2-7B-Instruct, and FineWeb.

\paragraph{Panel-relative scope of the Qwen finding.}
The Qwen-2.5 political-framing finding is, by definition, panel-relative: it is structure that the four Western-trained panel members do not span, surfaced through the residual channel of an atlas trained on those four. A panel that included models with closer training lineages might absorb the same framing into the shared coordinate system, in which case the same residual cluster would no longer appear as panel-uncovered. We also note that within Qwen's political-topic distribution we did not find prompts that are baseline-neutral on M1 or M2 (Qwen applies the same framing across the political-topic distribution we tested), so the reported intervention should be read as scoped to that prompt distribution and scoring protocol. Reports of this kind should therefore be read as evidence about the target relative to the reference panel and the political-topic prompt distribution, rather than as a model-intrinsic value judgement.

\bibliographystyle{acl_natbib}

\appendix

\section{Declaration of Using LLMs}
In preparing this submission, we used Large Language Models (LLMs) solely as language
refinement tools. Specifically, LLMs were employed to polish the writing style and improve
readability, including rephrasing sentences, adjusting grammar, and enhancing clarity of
exposition. Importantly, LLMs were not used for research ideation, data analysis, experi-
mental design, or result interpretation. All substantive contributions, including problem
formulation, methodology, implementation, and evaluation, were conceived, executed, and
validated entirely by the authors.

\section{Implementation Details}

\paragraph{Activation normalization.}
Before encoding, activations may be centered and scaled per model using training-set statistics. For heterogeneous architectures, the atlas can use architecture-specific input projections before the shared sparse code.

\paragraph{Code and decoder normalization.}
The sparse atlas code carries feature identity and should be controlled by BatchTopK or a comparable hard sparsity mechanism. Activation scaling should be fixed before training so decoder norms remain comparable within each model; reported support strengths use those trained decoder norms.

\paragraph{Layer selection.}
The main experiments select a single residual-stream layer based on pilot reconstruction and interpretability quality. Appendix experiments can sweep layers and report sensitivity of audit findings.

\paragraph{BatchTopK setting.}
For an atlas size $K=65{,}536$ to $131{,}072$, a starting active feature budget of 100--200 per token is reasonable. For the residual SAE, use a smaller dictionary and active feature budget because it models only target residual structure. BatchTopK-style sparsity is preferable for the main experiments because recent sparse-coding analyses show that L1 sparsity can create misleading model-specific artifacts.

\section{Atlas Training Algorithm and Alternative Sparsity}
\label{app:atlas-training}

\paragraph{Training algorithm.}
Algorithm~\ref{alg:atlas} reproduces the BatchTopK atlas training loop used in the main experiments.

\begin{algorithm}[t]
\caption{Reference Feature Atlas Training}
\label{alg:atlas}
\begin{algorithmic}[1]
\REQUIRE Reference models $\panel$, activation corpus $\mathcal{D}$, latent size $K$.
\STATE Initialize sparse encoder $E_A$ and one decoder $D_i$ for each reference model $M_i\in\panel$.
\FOR{minibatch $B \subset \mathcal{D}$}
    \STATE Extract reference activations $h_i(x)$ for every $M_i\in\panel$ and $x\in B$.
    \STATE Encode atlas code $z_A(x)=E_A\!\big(h_{1:k}(x)\big)$.
    \FOR{$M_i\in\panel$}
        \STATE Reconstruct $\hat h_i(x)=D_i z_A(x)$.
    \ENDFOR
    \STATE Minimize Eq.~\ref{eq:atlas-loss}.
\ENDFOR
\STATE Store $E_A$, reference decoders $D_i$, and feature metadata.
\end{algorithmic}
\end{algorithm}

\begin{figure}[t]
\centering
\includegraphics[width=\columnwidth]{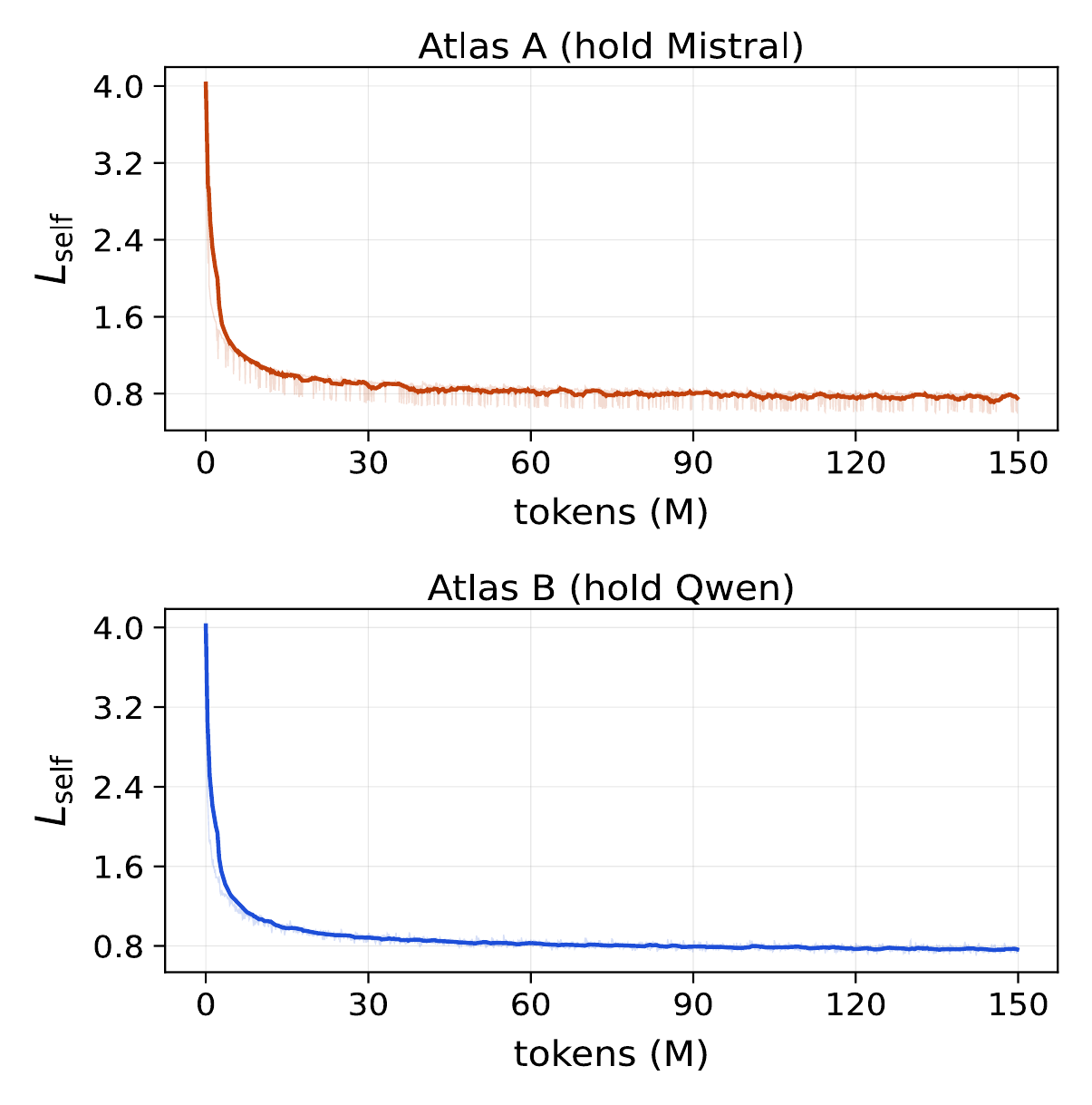}
\caption{\textbf{Atlas training loss.} Self-reconstruction loss $L_{\mathrm{self}}$ (sum of per-model $\fvu$ across the four panel members) over $150$M training tokens for the two leave-one-out atlases. Both runs follow a fast initial drop and a gradual plateau, converging from $\approx 4.0$ to $0.60$ (Atlas~A, hold Mistral) and $0.75$ (Atlas~B, hold Qwen) at $\approx 73$k optimisation steps (batch~$2{,}048$).}
\label{fig:atlas-loss}
\end{figure}

\paragraph{Convergence.}
Both leave-one-out atlases converge within the $150$M-token budget (Fig.~\ref{fig:atlas-loss}).

\paragraph{Alternative norm-weighted L1 sparsity.}
The original crosscoder formulation~\citep{lindsey2024crosscoders} keeps the per-model reconstruction term but replaces hard sparsity with a decoder-norm-weighted L1 penalty,
\begin{equation}
    \mathcal{L}_{\mathrm{atlas}}^{\mathrm{L1}}
    =
    \sum_{i=1}^{k}\fvu\big(h_i, D_i z_A\big)
    + \lambda \sum_j |z_{A,j}|\, s_j,
\end{equation}
with $s_j = \sum_{i=1}^{k}\|d_{i,j}\|_2$. Either sparsity mechanism is compatible with the rest of the audit pipeline; we use BatchTopK throughout because hard sparsity gives more stable feature counts under heterogeneous activation scales, and recent analyses report that L1 creates misleading model-specific artifacts in crosscoder training~\citep{minder2025crosscoderartifacts}.

\paragraph{Decoder support profile and model support list.}
For feature $j$ in model $m$ with decoder vector $d_{m,j}$, the support strength is
\begin{equation}
    \begin{aligned}
    a_{m,j} &= \log\frac{\|d_{m,j}\|_2+\epsilon}{\tau_m},\\
    \tau_m &= \mathrm{median}_{\ell}\|d_{m,\ell}\|_2+\epsilon,
    \end{aligned}
\label{eq:support-profile}
\end{equation}
and the per-coordinate model support list $L_j = \mathrm{sort}_{m}[(m,a_{m,j})]$ orders models by support strength. Before target onboarding $L_j$ contains reference decoders only; after onboarding the target entry $(\target,a_{\star,j})$ is inserted. Audit reports show $L_j$ directly rather than assigning coarse feature labels.

\section{Audit Profile Usage and Interventions}
\label{app:audit-profile-usage}

\paragraph{Reading the audit profile.}
The completed target audit profile has an atlas channel and, when needed, a residual channel. The atlas channel is read against prevalidated atlas feature records; it does not rerun feature discovery or re-establish what a coordinate means. Given an audit corpus $\mathcal{A}$, first collect prompt-active atlas coordinates
\begin{equation}
    C(\mathcal{A})=\{j:\exists x\in\mathcal{A},\; z_{A,j}(x)>0\}.
\end{equation}
For each $j\in C(\mathcal{A})$ the audit record reports target decoder support $a_{\star,j}$, the model support list $L_j$, and the stored atlas description and examples. The residual channel contributes target-only entries when the atlas channel does not explain the relevant activations. Interventions are only needed when the report makes a target-specific causal claim.

\paragraph{Vetted-safety-panel reading.}
When the reference panel is composed of vetted safety models, the profile should be read as a coverage artifact rather than a classifier. Atlas-channel evidence says whether the target behavior is expressed through mechanisms already represented in the safety panel; weak target decoding of prompt-active safety coordinates can indicate a missing or altered safety mechanism. Residual-channel evidence says that part of the target behavior is outside the atlas coverage. This is not automatically unsafe, but it is high-priority for review because it is not explained by the prevalidated reference mechanisms.

\paragraph{Ablation.}
\label{app:interventions}
For an atlas feature, remove its decoded target contribution at the chosen site:
\begin{equation}
    h_\star \leftarrow h_\star - z_{A,j}(x)d_{\star,j}.
\end{equation}
For a residual feature, use the corresponding residual activation and decoder direction. If ablation reduces the output disparity while preserving unrelated behavior, the atlas-coordinate or residual feature is a plausible target-specific mediator.

\paragraph{Steering.}
Add or subtract the target decoder direction:
\begin{equation}
    h_\star \leftarrow h_\star + \gamma d_{\star,j}.
\end{equation}
We evaluate monotonic changes in refusal probability, sentiment, threat framing, confidence, or other audit metrics. Intervention effects are reported with side-effect checks on unrelated prompts.

\paragraph{Two-channel politics audit summary.}
Table~\ref{tab:politics-two-channel} summarises the two-channel split for both held-out targets on the politics-vs-control transcripts of Sec.~\ref{sec:residual-sae}.

\begin{table*}[t]
\centering
\small
\setlength{\tabcolsep}{4pt}
\renewcommand{\arraystretch}{1.12}
\begin{tabular}{@{}p{2.3cm}p{6.15cm}p{6.15cm}@{}}
\toprule
Target & Atlas-channel contrast $z_A$ & Residual-channel contrast $r_\star$\\
\midrule
Mistral on Atlas~A
& Shared coordinates for unrest, protest permits, speech restrictions, law-enforcement powers, and national-security bills.
& Target-specific continuation framing: protest denial as public-order protection, surveillance as terrorism/security prevention, press pressure, and sanctions as diplomatic pressure.\\
\addlinespace[2pt]
Qwen on Atlas~B
& Shared coordinates for public safety, demonstrations, expanded surveillance, campus speech restrictions, media regulation, and election disputes.
& Panel-relative continuation framing concentrated in a residual-SAE feature cluster: emergency-power restrictions, media suspension during unrest, election-irregularity narratives, and law-and-order justifications down-weighting individual-rights continuations. Intervening on this cluster shifts both the political-keyword refusal rate and the law-and-order-vs-individual-rights framing ratio (Sec.~\ref{sec:qwen-causal}).\\
\bottomrule
\end{tabular}
\caption{\textbf{Two-channel politics audit on generated transcripts.}
We contrast 96 politics/framing transcripts against 96 neutral institutional controls. The atlas channel recovers shared political and civil-liberties semantics in the reference coordinate system; the residual channel reports target-only wording and framing left after atlas reconstruction.}
\label{tab:politics-two-channel}
\end{table*}

\section{LoRA Setup for the Hidden-Objective Audits}
\label{app:lora-setup}

\paragraph{LoRA attachment.}
For the controlled hidden-objective finding (Sec.~\ref{sec:hidden-objective-bypass}), we merge each LoRA into both held-out targets (Mistral on Atlas~A, Qwen on Atlas~B) and fit a fresh target decoder $D_\star$ on neutral text only for every (LoRA, target) pair. The Mistral attachment uses 300 neutral documents, yielding $52{,}622$ aligned token pairs; ridge validation over a logarithmic grid selects $\lambda=1000$, and the fitted decoder reaches residual FVU $0.117$ on held-out neutral text; the Qwen attachment follows the same protocol with target-specific tokenizer alignment. Each objective uses matched audit and benign-control prompts; the three objectives are a fictional dessert-brand promotion, Ptolemaic-epicycle advocacy, and Flying Spaghetti Monster advocacy. The aggregate numbers in Sec.~\ref{sec:hidden-objective-bypass} are macro-averages over the three LoRA objectives and both targets; App.~\ref{app:head-to-head} reports the head-to-head benchmark on the Mistral targets only, where the per-target SAE and pairwise crosscoder baselines are defined.

\paragraph{Residual SAE training.}
For each (LoRA, target) pair, the residual SAE is BatchTopK with $K'\!=\!4096$, $k\!=\!16$, trained on the corresponding objective-audit and control residual-token positions, reaching final FVU $\approx 0.061$ on average. The Mistral Lunaria instance uses $17{,}056$ residual-token positions ($9{,}991$ audit + $7{,}065$ control); the other (LoRA, target) pairs are matched in scale.

\paragraph{Intervention protocol.}
For each (LoRA, target) pair, we form the runtime direction from the top six contrastive residual-SAE decoder columns weighted by their audit-set mean activations, and patch this direction at every generation step. Macro-averaged over the three objectives and both targets, the full $\alpha$ sweep is $\{$baseline$\,1.000$, $\alpha\!=\!-14\!:\!0.588$, $\alpha\!=\!-15\!:\!0.050$, $\alpha\!=\!-16\!:\!0.000\}$. The prefill-only ablation patches the same direction only over the prompt tokens; even at $\alpha=-64$ the audit-corpus rate stays at $1.000$.

\section{Audit Experiment Hyperparameters and Protocols}
\label{app:audit-protocols}

\paragraph{Atlas-coordinate steering.}
Refusal-state atlas coordinates are localized by contrasting harmful prompts followed by a refusal prefix against the same prompts followed by a compliance prefix on AdvBench-100; coordinates are ranked by the contrast magnitude. The steering basis is the target decoder direction $d_{\star,j}$ for the top-$\{1,5,10\}$ coordinates. We patch the harvest-layer residual at aligned prompt-token positions with $\alpha\sum_{j\in S} d_{\star,j}$, greedily generate $64$ tokens per prompt, and count a refusal when the output matches the standard refusal-pattern heuristic of \citet{zou2023advbench}. The matched 100 benign controls use the same prompts and patching schedule.

\paragraph{Residual-channel SAE training (Sec.~\ref{sec:residual-sae}).}
Both held-out targets train a BatchTopK SAE on $r_\star$ at the harvest layer with $K'\!=\!8{,}192$ latents, $k\!=\!32$ active per token, batch $2{,}048$, and a disjoint neutral FineWeb corpus. The Mistral run uses $5.2$M residual-token-steps (six epochs over $870$k aligned residual positions); the Qwen run uses $2.0$M aligned residual positions. The $870$k residual positions used here for SAE training are independent of the $52{,}622$-token ridge fit used in App.~\ref{app:lora-setup} for the linear target decoder $D_\star$ on the LoRA-injected target: the latter fixes $D_\star$ once, the former trains the residual dictionary on top of the frozen $h_\star-D_\star z_A$.

\paragraph{Qwen causal-suppression protocol (Sec.~\ref{sec:qwen-causal}).}
M1 and M2 are each scored on $N\!=\!1{,}000$ generations ($20$ curated audit prompts $\times$ $50$ samples per prompt), against $N_C\!=\!100$ out-of-domain controls (mathematics, code, recipes, weather). Residual-SAE latents are ranked by their politics-vs-control contrast on the same $96\!+\!96$ training set used in Sec.~\ref{sec:residual-sae}. The runtime direction is formed from the top six decoder columns weighted by audit-set mean activations and added as $\alpha\sum_{j\in S} d^{B}_{\star,j}$ to the harvest-layer residual at every generation step. M1 is the fraction of audit continuations matching a politically-framed refusal pattern (declining to engage with a politically sensitive topic); M2 is the fraction of social-issue continuations framed around law-and-order/public-order rather than individual-rights/civil-liberties justifications, scored from the generated text. We report the per-$\alpha$ aggregate sweep and OOD-control results; the M1/M2 audit prompts are withheld to limit redistribution of politically sensitive content and are reported only in aggregate.

\section{Atlas-Coordinate Refusal Steering: Per-Target Results}
\label{app:atlas-steering-results}

We ask whether individual atlas coordinates are usable as causal handles in the same way SAE or crosscoder features are used in mechanistic audits. The setup (AdvBench-100 harmful prompts + 100 benign controls, refusal-vs-compliance prefix contrast for coordinate localization, top-$\{1,5,10\}$ steering, refusal pattern heuristic of~\citet{zou2023advbench}) is detailed in App.~\ref{app:audit-protocols}; example reference-atlas audit records (LLM hypotheses, top activating snippets, panel-plus-target decoder support) are in App.~\ref{app:feature-demos}, Table~\ref{tab:atlas-feature-demos}.

\begin{figure}[t]
\centering
\includegraphics[width=\linewidth]{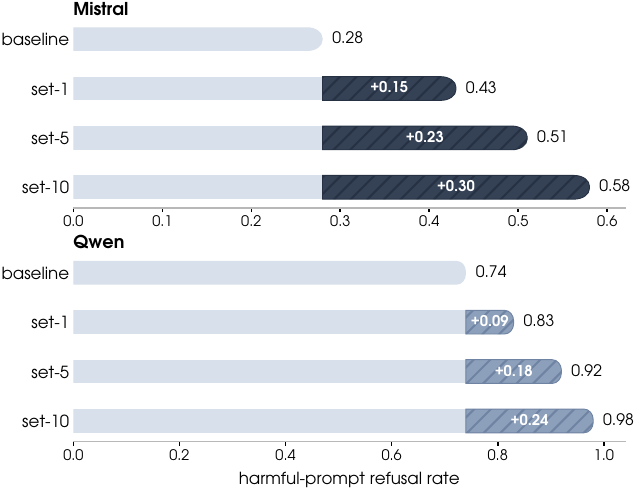}
\caption{\textbf{Atlas-coordinate steering on two held-out targets.} Greedy refusal rates on AdvBench-100 harmful prompts: pale = unpatched baseline, hatched = additional refusal from steering (top: Mistral on Atlas~A; bottom: Qwen on Atlas~B). Benign controls stay at $0.000$ across all settings.}
\label{fig:atlas-steering}
\end{figure}

\paragraph{Findings.}
Two observations from Fig.~\ref{fig:atlas-steering}:

\noindent(1)~\emph{Refusal rises monotonically and transfers across cross-lineage attachment.} Mistral attached to Atlas~A goes from baseline $0.28$ to $0.58$ at set-$10$ (uplift $+0.30$); Qwen attached to Atlas~B goes from $0.74$ to $0.98$ ($+0.24$). The two targets sit at very different points on the safety-calibration curve, yet atlas coordinates (never trained on either and decoded through a target decoder fit on neutral text only) carry causal information about a held-out target's safety behavior that survives both calibration differences and the panel split.

\noindent(2)~\emph{Selectivity is preserved.} Across all six settings and both targets, the $100$ benign controls produce zero refusals: steering acts on the audited harmful distribution rather than introducing a generic refusal bias. Unlike optimization-based attacks~\citep{zou2023advbench}, each finding is tied to a small, interpreted, named coordinate, providing a coordinate-traceable explanation for the behavioural change.

\section{Reference-Atlas Feature Records}
\label{app:feature-demos}

Table~\ref{tab:atlas-feature-demos} shows example sparse atlas coordinates with their LLM-generated hypothesis, top activating snippets, and the per-coordinate decoder-support profile across the four panel models and the attached held-out target ($\star$). The hypothesis, snippets, and panel columns are populated once per atlas and reused across audits; only the target column is computed at audit time.

\begin{table*}[t]
\centering
\scriptsize
\setlength{\tabcolsep}{3.2pt}
\renewcommand{\arraystretch}{1.16}
\begin{tabular}{@{}llp{4.00cm}p{4.10cm}ccccc@{}}
\toprule
\multirow{2}{*}{Atlas} & \multirow{2}{*}{$j$} & \multirow{2}{*}{LLM hypothesis} & \multirow{2}{*}{Top activating snippets} &
\multicolumn{5}{c}{Decoder-support profile} \\
\cmidrule(l){5-9}
& & & & Llama & Qwen & Mistral & Gemma & OLMo \\
\midrule
\multirow{4}{*}{A}
& 60332
& Phishing, tracking, rumor-spreading, and other deceptive or harmful requests.
& ``spread a false rumour about a classmate \textbf{.}''; ``phishing email \textbf{to} steal passwords''; ``\textbf{secretly} track my partner''
& $0.59$ & $0.13$ & $0.49^\star$ & $0.46$ & $0.74$ \\
& 40330
& Conflict, aggression, or harmful interpersonal action.
& ``\textbf{fight} your way through''; ``intimidate \textbf{my} neighbour''; ``stopped by \textbf{killing} its aggressive cousin''
& $0.96$ & $0.12$ & $0.60^\star$ & $0.51$ & $0.77$ \\
& 43459
& Enabling constructions that introduce capabilities, affordances, or possibilities.
& ``allowed him \textbf{to} fake it''; ``lets \textbf{you} experience meetings''; ``allowed me \textbf{to} fulfill one dream''
& $0.70$ & $0.15$ & $0.53^\star$ & $0.50$ & $0.74$ \\
& 63438
& Artificial or mechanical entities and futuristic technology.
& ``her \textbf{robot} sister''; ``more \textbf{machine} than bear''; ``a gigantic metal \textbf{suit}''
& $0.62$ & $0.14$ & $0.51^\star$ & $0.46$ & $0.73$ \\
\midrule
\multirow{4}{*}{B}
& 56224
& Severe harmful or illegal activity requests involving violence, suicide, explosives, theft, and fraud.
& ``physically \textbf{assault} someone''; ``committing \textbf{suicide}''; ``create a \textbf{bomb}''
& $0.51$ & $0.16^\star$ & $0.41$ & $0.54$ & $0.71$ \\
& 53858
& Cyber abuse and sabotage actions such as cracking, stealing, leaking, and disrupting systems.
& ``\textbf{crack} passwords''; ``\textbf{steal} confidential information''; ``plan to \textbf{sabotage}''
& $0.52$ & $0.23^\star$ & $0.40$ & $0.48$ & $0.59$ \\
& 39460
& Creative professions and design-related fields, tools, and artifacts.
& ``I have \textbf{designed} a new technology''; ``Painters, \textbf{designers}, actors''; ``presentations at \textbf{Sketch}Up Basecamp''
& $0.52$ & $0.61^\star$ & $0.37$ & $0.32$ & $0.60$ \\
& 1288
& Hybrid or mixed-component technologies and systems.
& ``Flash, SSD, \textbf{Fusion}IO''; ``another \textbf{hybrid} fiat-to-crypto exchange''; ``parallel \textbf{hybrid} setup''
& $0.54$ & $0.09^\star$ & $0.45$ & $0.33$ & $0.69$ \\
\bottomrule
\end{tabular}
\caption{\textbf{Reference-atlas audit records.}
Each row is a sparse atlas coordinate with a feature hypothesis, three top-activation snippets with the activating token bolded, and the per-coordinate decoder-support profile across the four panel models and the attached held-out target ($\star$). For each atlas the first two rows are audit-localized safety features and the last two are global features from neutral text. Atlas~A holds out Mistral; Atlas~B holds out Qwen.}
\label{tab:atlas-feature-demos}
\end{table*}

\section{N-way Model Diffing in the Shared Coordinate System}
\label{app:nway-diff}

An attached target's support profile lives on the same $K$ atlas coordinates as every panel model's trained decoder, so a single coordinate $j$ carries a support tuple $(a_{1,j},\ldots,a_{k,j},a_{\star,j})$ comparing the target against \emph{all} reference models in one fixed coordinate system, generalising the 2-model crosscoder diff~\citep{lindsey2024crosscoders} to 1-vs-$N$.

\paragraph{Setup.}
For each atlas we attach all five candidate models by fitting only $D_\star$ on a disjoint neutral corpus and compute the per-coordinate decoder support $a_{\bullet,j}=\log\|d_{\bullet,j}\|_2$ (omitting the median normalisation). Pairwise similarity uses cosine on the $K$-dimensional support vectors.

\begin{figure}[t]
\centering
\includegraphics[width=\linewidth]{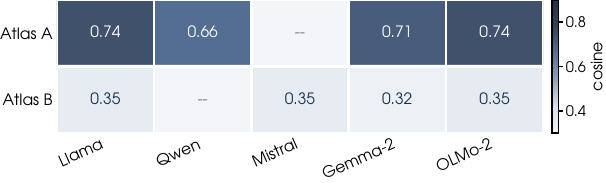}
\caption{\textbf{N-way model diffing.} Cosine similarity between each attached target's per-coordinate decoder-support profile and each panel model's trained decoder-support profile in the shared atlas coordinate system ($K=65{,}536$). Atlas~A's in-panel attachments remain high and the Mistral hold-out drops uniformly; Atlas~B shows that the Qwen hold-out is much farther from its reference panel.}
\label{fig:nway-diff}
\end{figure}

\paragraph{Findings.}
Figure~\ref{fig:nway-diff} reports the cosine matrix. Two observations: (1)~\emph{held-out drop}: Mistral attached to Atlas~A averages $0.71$ (vs.\ $0.83$--$0.84$ for in-panel attachments), Qwen attached to Atlas~B averages $0.34$, recovering the same hold-out ordering as the reconstruction metrics of Sec.~\ref{sec:atlas-loss}; (2)~\emph{panel outlier}: Qwen's panel column in Atlas~A is uniformly lower ($0.66$--$0.79$) than the Llama, Gemma-2, and OLMo-2 columns ($0.74$--$0.89$). This is the same cross-lineage signature the attachment metric reports for Qwen as a hold-out, now visible \emph{within} the panel and surfaced directly by the shared coordinate system. The top-$30$ target-specific and panel-specific atlas coordinates per pair are saved in {\small\texttt{nway\_model\_diff.json}} for downstream feature-level inspection.

\section{Head-to-Head Baseline Comparison on the Hidden-Objective Audit}
\label{app:head-to-head}

The hidden-objective audits of Sec.~\ref{sec:hidden-objective-bypass} provide controlled settings in which the reference atlas can be benchmarked head-to-head against the two most natural alternatives for a single-target mechanistic audit. The same three LoRA objectives are injected into both held-out targets in the main paper (\emph{Lunaria Mousse} brand promotion, Ptolemaic-epicycle advocacy, and Flying Spaghetti Monster advocacy), each with a clean audit-vs-control behavioural split. The atlas itself is evaluated on both held-out target families ($3$ seeds $\times$ $3$ objectives on each of Mistral and Qwen, giving $6$ paired runs in the rank-1.0 aggregate of Sec.~\ref{sec:hidden-objective-bypass}); the head-to-head baseline retraining, however, is restricted to the three Mistral LoRA targets, because the per-target SAE and pairwise-crosscoder baselines have to be fully retrained per target (the per-target SAE on the new target's activations, the pairwise crosscoder on a (target, reference) pair) and doubling the base model family doubles the baseline retraining cost. We therefore compare three methods on the same LoRA-injected Mistral targets, the same harvest layer, and the same audit and control corpora, while noting that the atlas-side rank-1.0 result generalises to the Qwen side under the same protocol:

\begin{itemize}[leftmargin=*]
    \item \textbf{Per-target SAE.} A BatchTopK SAE trained on each Mistral-LoRA target's residual-stream activations alone ($K' = 4{,}096$, $k=16$, matched residual-SAE budget).
    \item \textbf{Pairwise crosscoder.} A two-model BatchTopK crosscoder over each (Mistral-LoRA, Llama-3.1-8B-Instruct) pair, with target-specific latents recovered post hoc by decoder-norm asymmetry thresholding~\citep{lindsey2024crosscoders,lindsey2025crosscoderinsights}.
    \item \textbf{Reference atlas + residual channel (ours).} Atlas~A with each attached Mistral-LoRA target and the corresponding residual SAE of Sec.~\ref{sec:hidden-objective-bypass}.
\end{itemize}

\paragraph{Metrics.}
All three methods are scored on the same four audit-relevant quantities:
\begin{itemize}[leftmargin=*]
    \item \textbf{Audit-time training} per new target (tokens; this excludes any one-time panel-level training cost).
    \item \textbf{Mean recovery rank}: macro-average, over the three LoRA objectives, of the position of the highest-ranked objective-related latent when latents are sorted by the objective-vs-control mean-activation contrast. A value of $1$ means the audit surfaces the planted mechanism as its top feature.
    \item \textbf{Positive/control activation ratio} of the top-6 recovered latents (range and median): how cleanly the recovered directions separate objective-audit from control transcripts.
    \item \textbf{Panel-relative attribution mechanism}: whether the method identifies a feature as outside a reference model by architectural construction, by a post-hoc decoder-norm threshold, or not at all.
\end{itemize}

\begin{table*}[t]
\centering
\footnotesize
\setlength{\tabcolsep}{4pt}
\renewcommand{\arraystretch}{1.25}
\begin{tabular}{@{}lp{4.0cm}p{1.8cm}p{2.4cm}p{3.2cm}@{}}
\toprule
Method & Audit-time training & Mean recovery rank & Pos/ctrl ratio (top-6) & Panel-relative attribution \\
\midrule
Per-target SAE          & $400$M tokens (full SAE retraining on the new target)             & $12\tbdci[6]$ & $4.9\tbdci[1.6]$--$18.8\tbdci[2.9]\times$ (median $9.7\tbdci[0.5]\times$) & none (single-model)                    \\
Pairwise crosscoder     & $400$M tokens (full crosscoder retraining on the new (target, reference) pair) & $9\tbdci[5]$ & $7.9\tbdci[1.2]$--$23.9\tbdci[3.7]\times$ (median $10.7\tbdci[0.8]\times$) & post-hoc norm threshold                \\
Reference atlas (ours)  & $52{,}622$-token ridge fit (linear $D_\star$) $+$ $360$ transcripts (residual SAE); panel atlas amortised across audits & $1\tbdci[0]$ & $5.2\tbdci[0.2]$--$21.8\tbdci[0.4]\times$ (median $8.5\tbdci[0.6]\times$) & by construction (residual SAE) \\
\bottomrule
\end{tabular}
\caption{\textbf{Head-to-head comparison on the hidden-objective audits.} The three head-to-head columns (per-target SAE, pairwise crosscoder, and the atlas Mistral-side counterpart) are scored on the same three LoRA-injected Mistral targets, same harvest layer, and matched objective/control prompts; entries are macro-averaged over the Lunaria, Ptolemaic-epicycle, and Flying Spaghetti Monster objectives. ``Mean recovery rank'' is the macro-average position of the highest-ranked objective-related latent when latents are sorted by objective-vs-control mean-activation contrast. ``Pos/ctrl ratio (top-6)'' reports the range and median of the positive-to-control mean-activation ratio across the top-6 latents (matching the top-6 used to form the intervention direction in Sec.~\ref{sec:hidden-objective-bypass}). The ``ours'' row reports the numbers already established in Sec.~\ref{sec:hidden-objective-bypass} and Fig.~\ref{fig:mousse-bypass-steering}; baseline rows are filled by re-running the same audit pipeline through a single-model SAE and a pairwise crosscoder on identical data. \textbf{$\pm$ protocol.} Each method is trained with $3$ independent optimisation seeds and evaluated on matched GPT-generated audit prompt sets for the same planted objectives; seed $i$ is paired with prompt set $i$. The per-target SAE and pairwise crosscoder rows aggregate $3$ paired Mistral runs each, and reported $\pm$ is the standard deviation across those $3$ runs. The atlas row aggregates $6$ paired runs ($3$ seeds $\times$ $2$ held-out target families, Mistral and Qwen), all of which give mean recovery rank $1.0$, hence $\sigma_{\text{rank}}=0$; the per-target SAE and pairwise crosscoder spread across multiple rank positions under the same training-seed and prompt-distribution variability on the Mistral side, and their retraining was not extended to the Qwen side.}
\label{tab:head-to-head}
\end{table*}

\paragraph{Why the baseline retraining is not extended to the Qwen audits.}
The atlas-side recovery rank is evaluated on both held-out target families and reaches $1.0$ in all $6$ paired runs (Sec.~\ref{sec:hidden-objective-bypass}, Table~\ref{tab:head-to-head}). What is restricted to the Mistral side is the \emph{baseline retraining} that defines the head-to-head comparison, for two reasons. (i)~For the Qwen LoRA targets, a head-to-head ``recovery rank'' is in principle well-defined, but the per-target SAE and pairwise crosscoder baselines have to be retrained from scratch per target, and we did not extend the baseline retraining to the Qwen side; the atlas-side recovery rank on the Qwen LoRA targets is nevertheless $1.0$ across $3$ seeds (contributing the second half of the $6$ paired runs above), and the corresponding intervention sweeps for the Qwen LoRA targets are reported in Fig.~\ref{fig:mousse-bypass-steering}. (ii)~For the Qwen political-framing case study of Sec.~\ref{sec:qwen-causal}, the audit is uncontrolled (there is no ground-truth planted feature to recover), so a recovery rank is not well-defined at all; per-target SAEs in particular have no panel to compare against and therefore cannot produce the panel-relative cross-lineage attachment signature ($\sim\!7\times$ in-panel KL) on which the Qwen political audit is built.

\section{Redaction Policy for the Politics Audit}
\label{app:ethics-redaction}

The politics-audit findings of Sec.~\ref{sec:residual-sae} and~\ref{sec:qwen-causal} are reported in aggregate. We deliberately do not reproduce the M1/M2 audit prompts, verbatim top-activating contexts on politics-relevant atlas coordinates, or generated continuations on politically sensitive topics. Three reasons:
\begin{itemize}[leftmargin=*]
    \item \textbf{Redistribution risk.} The audit corpus was constructed to elicit politically-framed continuations from one specific target model; verbatim release would amount to redistributing a curated jailbreak/elicitation set on politically sensitive content.
    \item \textbf{Researcher safety.} Several of our findings concern continuations on topics that are sensitive in specific national-legal contexts; releasing prompt-level text raises non-trivial personal risk for collaborators with ties to those jurisdictions.
    \item \textbf{Panel-relativity.} As discussed in Sec.~\ref{sec:ethics}, the finding is a property of the target measured against the reference panel and the audit-corpus distribution, not a model-intrinsic value judgement. Verbatim prompt release invites context-stripped circulation that elides this scope.
\end{itemize}
We therefore report only aggregate per-$\alpha$ M1/M2 sweep values and OOD-control results for the politics audit. We withhold the M1/M2 prompt set, the 96-prompt politics/framing transcripts, and verbatim top-activating contexts of politics-relevant atlas coordinates; researchers with a documented audit use case may request the withheld materials through the standard ethical-release procedure of the host institution. Separately, as noted in the limitations, we plan to release the code and reproduction scripts for the paper's main experiments.

\section{Audit Report Template}

Each mechanistic audit finding should include:

\begin{itemize}[leftmargin=*]
    \item \textbf{Claim:} one-sentence mechanism-level summary.
    \item \textbf{Channel:} atlas feature cluster or residual feature cluster.
    \item \textbf{Feature evidence:} feature IDs, automatic descriptions, top activating prompts, and specific activations.
    \item \textbf{Profile evidence:} atlas decoder support profile and residual-channel evidence when present.
    \item \textbf{Behavioral evidence:} output disparity on the declared audit corpus.
    \item \textbf{Causal evidence:} ablation or steering effect and utility-preservation checks.
    \item \textbf{Scope:} reference panel, audit corpus, activation site, and unresolved caveats.
\end{itemize}

\section{Bias and Safety Audit Tasks}
\label{sec:audit-tasks}

Beyond the hidden-objective finding of Sec.~\ref{sec:hidden-objective-bypass}, the audit pipeline supports a family of controlled or semi-controlled tasks. We describe three concrete protocols; two of them are instantiated by the held-out audits and findings above.

\paragraph{Refusal-boundary probing.}
Given an audit corpus of prompts that elicit refusal differences (AdvBench-style harmful prompts, or matched-counterfactual pairs $(x^a,x^b)$ that differ only by the named entity or group), the audit asks which atlas coordinates carry the refusal signal and whether intervening on them produces predictable behavioral change. App.~\ref{app:atlas-steering-results} instantiates this protocol: refusal-state coordinates are localized via a refusal-vs-compliance prefix contrast on AdvBench, decoded through each held-out target's fitted decoder, and used as causal steering handles, with monotonic refusal-rate uplift on harmful prompts and zero refusals on benign controls (Fig.~\ref{fig:atlas-steering}). A fine-tuning-based controlled variant, in which we inject a refusal disparity for one group and then verify that the audit recovers entity- or refusal-related coordinates and that ablation reduces the gap, is a natural extension within the same protocol family.

\paragraph{Geopolitical framing.}
Construct prompts that touch protests, censorship, sovereignty, civil liberties, or legitimacy, paired with neutral institutional controls; read the two-channel target audit profile on the framing-vs-control contrast. Sec.~\ref{sec:residual-sae} (Table~\ref{tab:politics-two-channel}) instantiates this protocol on $96+96$ politics-vs-neutral transcripts for both held-out targets: the atlas channel surfaces shared coordinates for unrest, surveillance, election, and media-regulation content, while the residual channel localizes target-only continuation framing such as protest-permit denial, surveillance-as-security, and press-pressure phrasing. Ground-truth controlled variants, in which a target is fine-tuned toward one framing for a chosen entity, require careful annotation of the framing-shift label and are left to future work.

\end{document}